\def\year{2022}\relax
\documentclass[letterpaper]{article} 
\usepackage{aaai22}  
\usepackage{times}  
\usepackage{helvet}  
\usepackage{courier}  
\usepackage[hyphens]{url}  
\usepackage{graphicx} 
\usepackage{natbib}  
\usepackage{caption} 
\DeclareCaptionStyle{ruled}{labelfont=normalfont,labelsep=colon,strut=off} 
\usepackage{algorithm}
\usepackage{algorithmic}

\usepackage{newfloat}
\usepackage{listings}
\floatstyle{ruled}
\newfloat{listing}{tb}{lst}{}
\floatname{listing}{Listing}

\usepackage{multirow}
\usepackage{amsmath}
\usepackage{subcaption}
\usepackage{booktabs}
\usepackage{tabularx}
\newcolumntype{C}{>{\centering\arraybackslash}X} 
\usepackage{adjustbox}

\title{SimIPU: Simple 2D Image and 3D Point Cloud Unsupervised Pre-Training \\
for Spatial-Aware Visual Representations}
\author{
    Zhenyu Li,\textsuperscript{\rm 1}
    Zehui Chen,\textsuperscript{\rm 2}
    Ang Li,\textsuperscript{\rm 3}
    Liangji Fang,\textsuperscript{\rm 3}
    Qinhong Jiang,\textsuperscript{\rm 3} \\ 
    Xianming Liu,\textsuperscript{\rm 1}
    Junjun Jiang,\textsuperscript{\rm 1}\thanks{Corresponding Author}
    Bolei Zhou,\textsuperscript{\rm 4}
    Hang Zhao\textsuperscript{5}
}
\affiliations{
    \textsuperscript{\rm 1} Harbin Institute of Technology, 
    \textsuperscript{\rm 2} University of Science and Technology,\\
    \textsuperscript{\rm 3} SenseTime Research, 
    \textsuperscript{\rm 4} The Chinese University of Hong Kong, 
    \textsuperscript{\rm 5} IIIS, Tsinghua University\\

    \{zhenyuli17, csxm, jiangjunjun\}@hit.edu.cn, lovesnow@mail.ustc.edu.cn,\\
    \{liang1, fangliangji, jiangqinhong\}@senseauto.com, bzhou@ie.cuhk.edu.hk, 
    hangzhao@mail.tsinghua.edu.cn
}

\begin{document}

\maketitle

\begin{abstract}
    Pre-training has become a standard paradigm in many computer vision tasks. However, most of the methods are generally designed on the RGB image domain. Due to the discrepancy between the two-dimensional image plane and the three-dimensional space, such pre-trained models fail to perceive spatial information and serve as sub-optimal solutions for 3D-related tasks. To bridge this gap, we aim to learn a spatial-aware visual representation that can describe the three-dimensional space and is more suitable and effective for these tasks. To leverage point clouds, which are much more superior in providing spatial information compared to images, we propose a simple yet effective 2D Image and 3D Point cloud Unsupervised pre-training strategy, called~\textbf{SimIPU}. Specifically, we develop a multi-modal contrastive learning framework that consists of an intra-modal spatial perception module to learn a spatial-aware representation from point clouds and an inter-modal feature interaction module to transfer the capability of perceiving spatial information from the point cloud encoder to the image encoder, respectively. Positive pairs for contrastive losses are established by the matching algorithm and the projection matrix. The whole framework is trained in an unsupervised end-to-end fashion. To the best of our knowledge, this is the first study to explore contrastive learning pre-training strategies for outdoor multi-modal datasets, containing paired camera images and LIDAR point clouds. Codes and models are available at \url{https://github.com/zhyever/SimIPU}.
\end{abstract}

\section{Introduction}
\label{sec:intro}

Large-scale models have achieved significant success in deep learning, where fine-tuning after pre-training has become a well-established and commonly used paradigm, such as ELMo~\cite{peters2018deep}, GPT~\cite{brown2020language}, and BERT~\cite{devlin2018bert} in NLP. As for computer vision, benefiting from the massive amount of labeled data, the supervised pre-trained models on ImageNet~\cite{deng2009imagenet} have long dominated. In recent years, unsupervised pre-training strategies have drawn much more attention. Various successful methods~\cite{he2020momentum, chen2020simple, grill2020bootstrap} have achieved comparable or better results compared to supervised pre-trained ones on a few 2D tasks, including image classification, object detection, and image segmentation~\cite{jaiswal2021survey}. However, all these methods are designed on the two-dimensional image plane, which exists a large discrepancy between the three-dimensional space. As a result, such pre-trained models can not perceive spatial information and demonstrate limited performance improvement on 3D-related downstream tasks. Therefore, learning a spatial-aware representation that can describe three-dimensional space is much more essential.

\begin{figure}[t]
    \centering
    \includegraphics[width=3.2in]{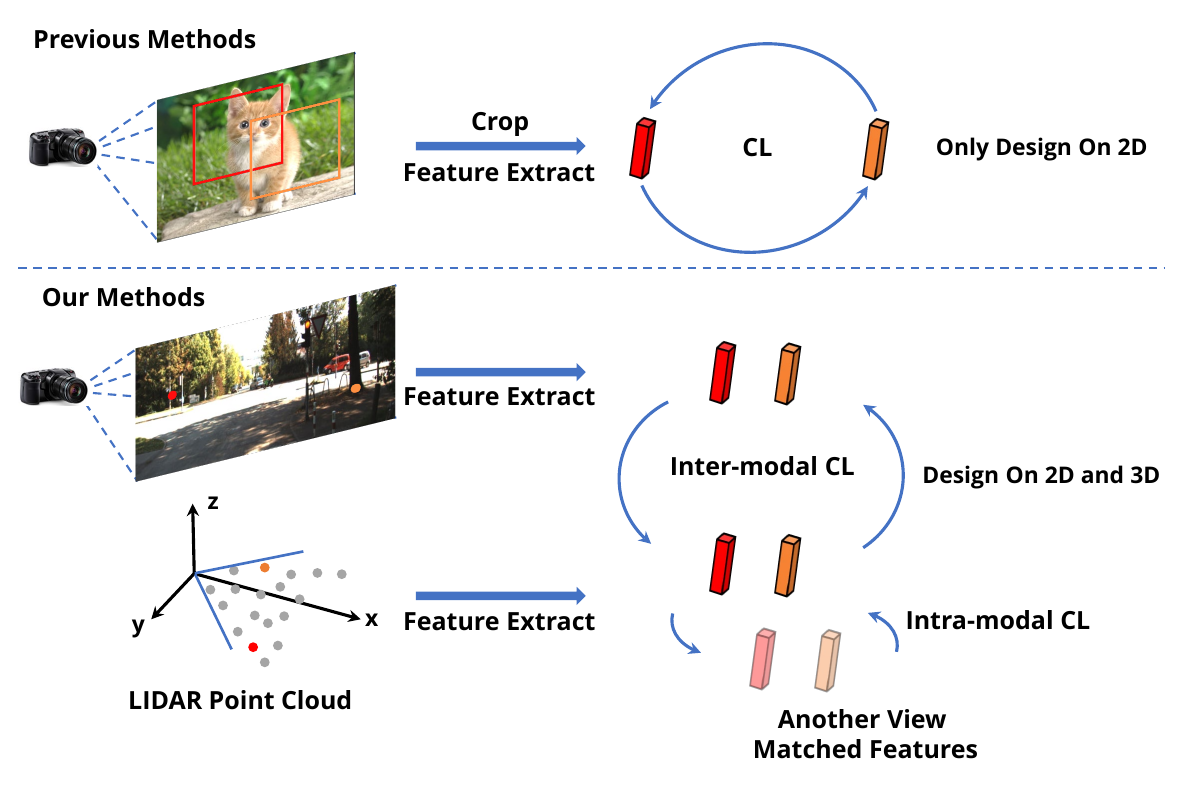}
    \caption{\textbf{Motivation}. SimIPU is designed on both the 2D image plane and the 3D space. Intra-modal module learns the spatial-aware representation with point clouds. Inter-modal module transfers the capability of extracting spatial-aware representations to the image-feature extractor. `CL' is the abbreviation of the contrastive learning.}
    \label{fig::teaser}
\end{figure}

Compared to images, point clouds are much more superior in providing spatial information~\cite{qi2017pointnet++}, which may lead them to be more suitable for learning such representations. PointContrast~\cite{xie2020pointcontrast} is the first study to explore pre-training strategies for point clouds. They utilize different scene views to generate positive pairs and adopt a PointInfoNCE loss to learn useful dense/local representations. Motivated by the success of 2D image and 3D point cloud pre-training, Prior3D~\cite{hou2021pri3d} propose a geometric prior contrastive loss to imbue the prior 3D information to image representations. However, there is no intra-modal constraint on point clouds, which can lead to trivial solutions. Furthermore, all of these methods focus on indoor RGB-D data, where point clouds are reconstructed by the depth value. As for the outdoor scene, the point cloud provided by LIDAR contains more noise and massive background points. It also lacks point-to-point correspondences, which makes the design of pre-training methods tougher.

In this paper, we develop a simple yet effective 2D image and 3D point cloud unsupervised pre-training framework for outdoor multi-modal data~(\textit{i.e.,} paired images and LIDAR point clouds) to learn spatial-aware visual representations. To solve the aforementioned problems, our method explicitly imposes the contrastive loss on point-cloud features to guarantee models to learn spatial-aware representations. We harness more robust and informative global features and apply the Hungarian algorithm and the projection matrix to associate the matching correspondences. To the best of our knowledge, this is the first study to explore pre-training strategies for outdoor multi-modal data.

Specifically, the framework consists of an intra-modal spatial perception module and an inter-modal feature interaction module. In terms of the spatial perception module, we adopt an intra-modal contrastive learning method to learn spatial-aware representations with point clouds. We utilize global transformation to yield two views of a point cloud and adopt an encoder to extract the global features. Then, the Hungarian algorithm is applied to establish the matching correspondences between the downsampled points. A contrastive loss serves to push the distances of matched point features closer. The certain equivariance learned from random geometric transformation leads to a spatial-aware representation. As for the feature interaction module, we adopt a similar contrastive learning strategy to transfer prior spatial knowledge from the point-cloud encoder to the image encoder. Therefore, positive pairs between image features and point-cloud global features are established through the projection process from LIDAR to the camera. Since both features are global representations, alignment is achieved on the fly. Benefiting from the inter-modal interaction, the image encoder can gradually acquire the capability of extracting spatial-aware representations. In the pre-training stage, the whole framework is trained in an unsupervised end-to-end fashion. Our contributions are summarized as follows:

\begin{itemize}

\item We propose a simple yet effective pre-training method, termed~\textbf{SimIPU}. It exerts the advantages of massive unlabeled multi-modal data to learn spatial-aware visual representations that can further improve the model performance on downstream tasks. To the best of our knowledge, this is the first study to explore contrastive learning pre-training strategies for outdoor multi-modal data.

\item We develop a Multi-Modal Contrastive Learning framework, which consists of an intra-modal spatial perception module and an inter-modal feature interaction module. 

\item Our method significantly outperforms other pre-training counterparts when transferring the models to 3D-related downstream tasks, including 3D object detection (2.3\% AP), monocular depth estimation (0.12m RMSE) and monocular 3D object detection (0.6\% AP).

\end{itemize}

\section{Related Work}
\label{sec:related_work}

\begin{figure*}[t]
    \centering
    \begin{subfigure}[b]{0.32\linewidth}
        \centering
        \includegraphics[width=2.2in]{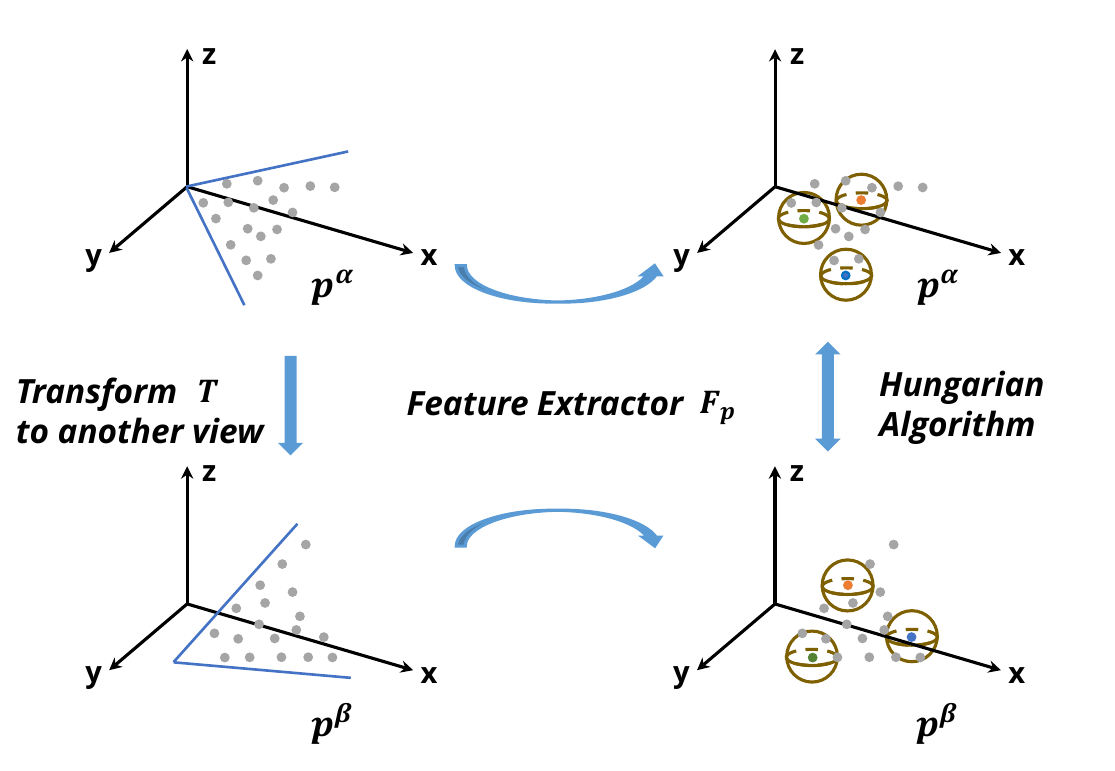}
        \caption{Spatial Perception Module}
        \label{fig::intro-modality::framework}
    \end{subfigure}
    \hfill
    \begin{subfigure}[b]{0.32\linewidth}
        \centering
        \includegraphics[width=2.2in]{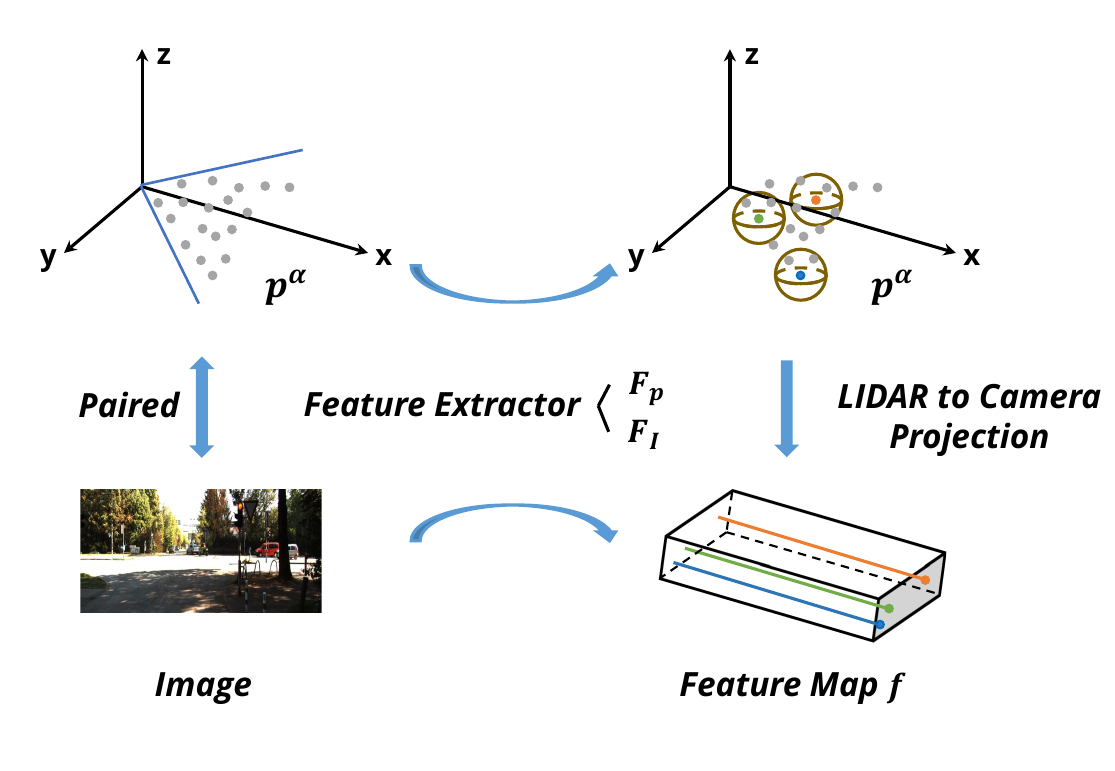}
        \caption{Feature Interaction Module}
        \label{fig::cross-modality::framework}
    \end{subfigure}
    \hfill
    \begin{subfigure}[b]{0.32\linewidth}
        \centering
        \includegraphics[width=2.2in]{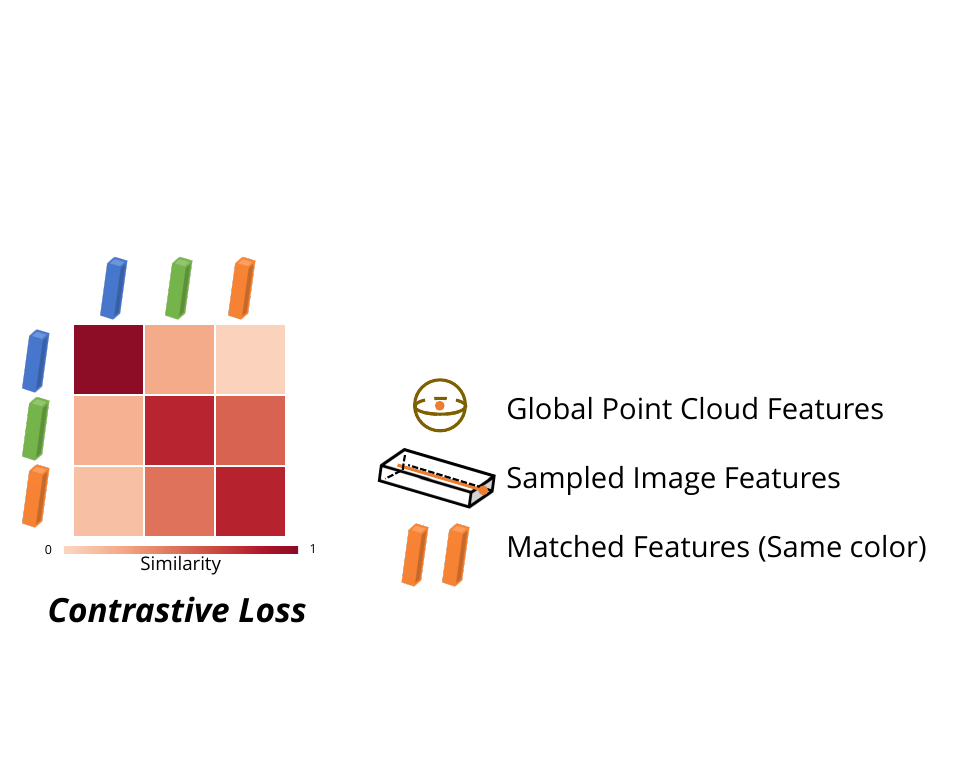}
        \caption{Contrastive Losses}
        \label{fig::losses::framework}
    \end{subfigure}
    \caption{\textbf{Framework of SimIPU}. Matched pairs are in the same color. The whole framework is trained in an end-to-end manner. (a) \textbf{Intra-Modal Spatial Perception Module}: We utilize set abstraction layers to extract global point cloud features and downsample points (results are in color) from different views. The Hungarian Algorithm is applied to match the downsampled points according to locations. (b) \textbf{Inter-Modal Feature Interaction Module}: We adopt a standard ResNet-50 to extract global image features. Projection matrix from point cloud to image plane establish the association between positive pairs. (c) \textbf{Contrastive Loss}: Contrastive losses are applied to push closer the distances of matched pair features.}
    \label{fig::framework}
\end{figure*}

\subsection{2D Self-supervised Representation Learning}
\label{subsec:2dcl}
Pretext task and contrastive learning are two key points of 2D self-supervised representation learning. There is a wide range of tasks that have been designed to learn useful visual representations, including colorization ~\cite{zhang2016colorful}, inpainting~\cite{pathak2016context}, spatial jigsaw puzzles~\cite{noroozi2016unsupervised}, and discriminate orientation~\cite{gidaris2018unsupervised}. Although the improvement is limited, these methods provide a possibility to achieve performance gains from pre-training strategies. SimCLR and SimCLR v2~\cite{chen2020simple} makes a breakthrough. They groundbreakingly proposed a discrimination pretext task. Contrastive loss is applied for pushing away the feature distances of different instances. MoCo and its improved version MoCo v2~\cite{he2020momentum} further utilize a memory bank to alleviate the constraints on large batch size. Beyond contrastive learning, BYOL~\cite{grill2020bootstrap} relies only on positive pairs, but it does not collapse in case a momentum encoder is used. All these methods are designed on the image plane and can be sub-optimal solutions for 3D-related downstream tasks. To circumvent the need for spatial-aware representations, some methods propose to learn representations from videos by using ego-motion as supervisory signal~\cite{jayaraman2015learning, agrawal2015learning, Lee_2019_BMVC} and self-supervised depth estimation~\cite{jiang2018self}. In this paper, we aim to further explore contrastive pre-training strategies following the successful trend of contrastive learning.

\subsection{3D Self-supervised Representation Learning}
Inspired by the success of 2D self-supervised representation learning, PointContrast~\cite{xie2020pointcontrast} introduces a contrastive pretext task in a 3D paradigm. Driven by indoor point cloud data properties, the same points in different frames compose positive pairs and are used for contrastive learning. To make full use of the point cloud data, Contrast Context~\cite{hou2021exploring} proposes to adopt a ShapeContext descriptor to divide the scene, which provides more negative pairs for contrastive learning and improve the effectiveness of pre-training models. CoCoNets~\cite{lal2021coconets} further explores self-supervised learning of amodal 3D feature representations agnostic to object and scene semantic content. The above methods focus on indoor RGB-D data. As for outdoor LIDAR point clouds, Pillar-Motion~\cite{luo2021self} propose a self-supervised pillar representation learning method that makes use of the optical flow extracted from camera images. Since the LIDAR point clouds lack point-to-point correspondences, there are fewer contrastive learning methods for pre-training.

\subsection{Multi-modal Representation Learning}
Much effort has been made into multi-modal representation learning. Based on paired image and text, \cite{yuan2021multimodal} propose a unified multi-modality contrastive learning framework to learn useful visual representations. Motivated by the success of 2D image and 3D point cloud pre-training, Pri3D~\cite{hou2021pri3d} further explore multi-modal pre-training methods to enhance the visual representation learning with indoor RGB-D data. Dense/local representations are learned, which boosts the performance of downstream tasks. However, there is no intra-modal constraint on point clouds, which can lead to trivial solutions and limit performance improvement. On the contrary, \cite{liu2021learning} propose a method to imbue the image prior to the 3D representation. All these methods motivate us to further explore multi-modal pre-training strategies for more challenging outdoor multi-modal data.

\subsection{3D Visual Tasks}
\label{subsec:3ddowntream}
We utilize three 3D visual tasks to evaluate the effectiveness of our method, which are fusion-based 3D object detection, monocular depth estimation, and monocular 3D object detection.  Fusion-based 3D object detection methods~\cite{zhang2020multi, sindagi2019mvx, liang2019multi, wang2020multi} combine the image and point cloud modality and learn the interaction between them. By exert multi-modal data, they achieve satisfying results compared to methods that only utilize point cloud data. In this paper, we mainly focus on this task and design extensive experiments to show the effectiveness of our method. For monocular depth estimation and monocular 3D object detection, which are two challenging 3D-related visual tasks on a single image modality, many methods~\cite{eigen2015predicting, bhat2021adabins, lyu2020hr, wang2021fcos3d} are proposed to improve the model performance. We execute experiments on these two tasks, which only utilize the single image modality to further evaluate the generalization of SimIPU.

\section{Method}
\label{sec:method}
In this section, we introduce our self-supervised pre-training pipeline in detail. First, to motivate the necessity of this multi-modal method, we conduct a pilot study to determine what kind of pre-training strategy we need in the downstream task that we are mainly focusing on: fusion-based 3D objection detection (Section \ref{subsection::method::pilot_study}). Then, we introduce our multi-modal self-supervised pre-training framework, including an Intra-Modal Spatial Perception module (Section \ref{subsection::method::intro}), and an Inter-Modal Feature Interaction module (Section \ref{subsection::method::cross}).  The overview of the proposed framework is shown in Fig.~\ref{fig::framework}.

\subsection{Pilot Study: Is 2D pre-training Useful?}
\label{subsection::method::pilot_study}
Previous fusion-based 3D objection detection methods utilize different kinds of pre-trained 2D feature extractors, including scratch models, ImageNet supervised classification pretrained models in ~\cite{chen2017multi, liang2019multi, wang2020multi}, and 2D detection pre-trained models in~\cite{sindagi2019mvx}, to initialize the backbone. However, there has been no discussion about which pre-training strategy can further improve the model performance on 3D object detection. To fill this blank, we execute this pilot study to assess the effect of different pre-trained models on fusion-based 3D object detection.

\begin{figure}[t]
    \centering
    \includegraphics[width=2.2in]{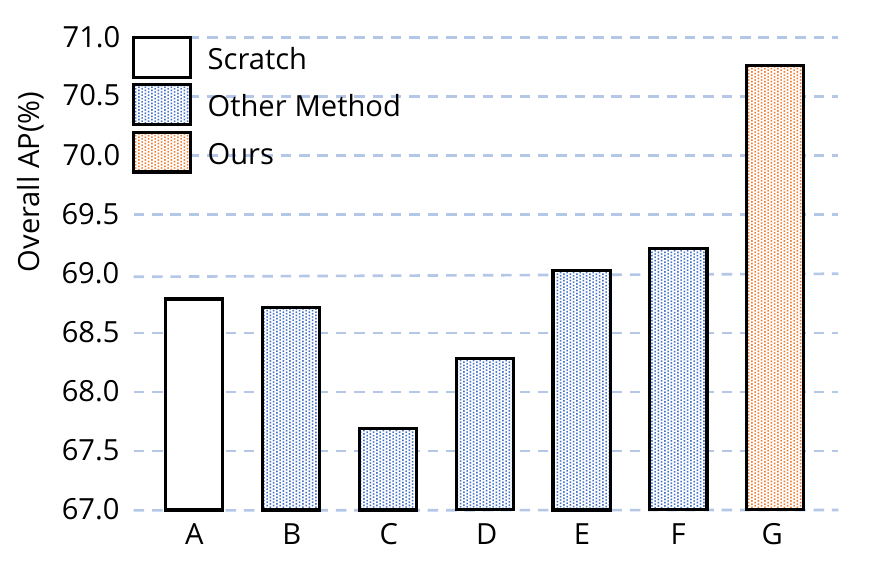}
    \caption{Pilot Study Results. In KITTI 3D object detection experiments, we adopt different pre-trained models, such as A. Scratch, B. 2D detection on CityScapes, C. 2D detection on KITTI, D. Supervised pre-trained on ImageNet, E. Moco-v2 on ImageNet, F. DenseCL on ImageNet, and G. Our method on KITTI, to initialize the backbones.}
    \label{fig::pilotstudy}
\end{figure}

We adopt the state-of-the-art fusion-based 3D object detection method MVXNet~\cite{sindagi2019mvx} with Moca~\cite{zhang2020multi} as our baseline, and train models on the KITTI~\cite{geiger2012we} dataset. We only change different pre-trained 2D feature extractor weights when initialization and keep all the other training settings the same. The results are shown in Fig.~\ref{fig::pilotstudy}. Critically, one can observe that pre-training models cannot improve the performance of the downstream task. These results suggest that the discrepancy between the two-dimensional image plane and three-dimensional space exists. All these pre-trained methods are designed on the 2D domain, which leads to sub-optimal solutions for the fusion-based 3D task.

Is there any way to learn a spatial-aware visual representation that is beneficial to 3D tasks? Without any label, it is extremely tough to directly learn such representations with the single image modality. Driven by massive multi-modal data, we can achieve it via multi-modal contrastive learning in an indirect manner. Specifically, we propose an intra-modal spatial perception module to learn a spatial-aware representation from point clouds and an inter-modal feature interaction module to transfer the capability of space perception to the image encoder.

\subsection{Intra-Modal Spatial Perception Module}
\label{subsection::method::intro}
We design the intra-modal contrastive learning module to pre-train a spatial-aware global representation with point clouds. The framework is shown in Fig.~\ref{fig::intro-modality::framework}. A key observation is that features at the same location in different views should be similar. 

To yield two different views of a point cloud, we sample a random 3D geometric transformation $T$ to  transform a given point cloud $p^{\alpha} \in R^{n\times c}$ into another view $p^{\beta} \in R^{n\times c}$:

\begin{equation}
\label{eq::transform}
p^{\beta} = T(p^{\alpha}),
\end{equation}
where $n$ is the point number in a scene, and $c$ is the channel of raw point features, which normally includes the 3D location and reflection rate. The superscript $\alpha$ and $\beta$ indicate two different views. In this work, we mainly consider the rigid transformation $T$, including rotation, translation, and scaling.

On completion of constituting two different views, we extract the global point cloud features by a PointNet++~\cite{qi2017pointnet++} encoder. Specifically, we apply several set abstraction layers for downsampling and extracting global context representations, which can be mathematically written as:
\begin{equation}
\label{eq::extract_features}
l^{\alpha}, f^{\alpha} = F_{P}(p^{\alpha}),\ l^{\beta}, f^{\beta} = F_{P}(p^{\beta}),
\end{equation}
where $l$ and $f$ are the location and feature of downsampled points, respectively. $F_{P}$ is the point cloud feature extractor. Note that our method is different from pre-training methods for indoor point clouds, which mainly focus on dense/local representations~\cite{xie2020pointcontrast, hou2021exploring, hou2021pri3d}. Since the outdoor data contains much more noise and massive background points, we utilize the global features to enhance the quality of the extracted representations. Therefore, during the feature extraction, meaningless information can be gradually filtered by the random sample strategy and the set abstraction layers. As a result, the spatial-aware representations will be well preserved, which results in certain spatial prior knowledge to transfer to the image encoder. Furthermore, random sampling can make the same point cloud generate different sampled points, which can improve the utilization of point cloud data and the effectiveness of the representation learning. 

However, extracting global features can introduce random properties, which leads to the inevitable mismatching of downsampled points. To construct positive pairs for contrastive learning, we utilize the Hungarian algorithm to achieve the positive correspondence matching $M_1$. The cost matrix of the bipartite assigning algorithm is computed by the $l_2$-norm distance between the downsampled points in two different views:
\begin{equation}
\label{eq::matching}
M_1 = assign(cost = distance(T(l^{\alpha}),\ l^{\beta})).
\end{equation}

Here, we apply the same transformation $T$ in Eq.\ref{eq::transform} to align the coordinates for the distance computation. The Hungarian algorithm can guarantee a favorable global-optimal matching and solves the problem of the lack of correspondences in outdoor multi-modal data.

After establishing the correspondence, we adopt the constrastive loss to push in the distance between the features of matched points. As for each matched pair $(i,\ j)\in M_1$, point feature $f_{i}^{\alpha}$ will serve as the query and $f_{j}^{\beta}$ will serve as the positive key $k^+$. We treat point feature $f_{k}^{\beta}$ where $(\cdot,\ k)\in M_1$ and $k \neq j$ as negative keys. We calculate the intra-modal contrastive loss as:
\begin{equation}
\label{eq::intro_loss}
L_{intra} = -\sum\limits_{(i,\ j)\in M_1}\log\frac{\exp{(f_{i}^{\alpha} \cdot f_{j}^{\beta}/\tau)}}{\sum_{(\cdot,\ k)\in M_1}{\exp{(f_{i}^{\alpha} \cdot f_{k}^{\beta}/\tau)}}},
\end{equation}
where $\tau$ is the temperature factor.

The intra-modal contrastive loss can push in the feature distance on a similar location of different views. Such a certain equivariance learned from random geometric transformation engenders a spatial-aware representation. 

\subsection{Inter-Modal Feature Interaction Module}
\label{subsection::method::cross}
To enable the image encoder the capability of perceiving spatial space, we propose the Inter-Modal Feature Interaction module, where the image feature extractor can gradually learn spatial-aware representations by embracing the inter-modal interaction. The framework is shown in Fig.~\ref{fig::cross-modality::framework}. 

Following most of the contrastive learning methods, we adopt a standard ResNet-50 as the default image backbone to extract global feature maps from given images:
\begin{equation}
\label{eq::extract_img_features}
f = F_{I}(I),
\end{equation}
where $f$, $F_{I}$, $I$ are the feature map, image feature extractor, and the input image, respectively.

\begin{table*}[t]
    \centering
    \begin{tabularx}{0.9\linewidth}{@{}c|*{3}{C}|*{3}{C}|*{3}{C}|*{2}{C}c@{}}
        \hline
        \multirow{2}{*}{Pre-train}  & \multicolumn{3}{c|}{Car $\text{AP}_{3D}$(\%)} & \multicolumn{3}{c|}{Pedestrian $\text{AP}_{3D}$(\%)}  & \multicolumn{3}{c|}{Cyclist $\text{AP}_{3D}$(\%)} & \multicolumn{3}{c}{Overall $\text{AP}_{3D}$(\%)}
        \\
        & Easy     & Mod.     & Hard   & Easy     & Mod.     & Hard   & Easy     & Mod.     & Hard   
        & Easy     & Mod.     & Hard 
        \\ \hline
        Scratch        & 86.18 & 76.57 & 74.08 
        & 67.95 & 62.18 & 57.24 
        & 83.37 & 66.99 & 63.11 
        & 79.17 & 68.58 & 64.81
        \\
        Ours-K         & 87.87 & 77.36 & 74.30 
        & 71.25 & 66.18 & 60.24 
        & 84.83 & 69.11 & 64.04 
        & \textbf{81.32} & \textbf{70.88} & \textbf{66.19}
        \\
        \hline
        Gain           & +1.69 & +0.79 & +0.22 
        & +3.30 & +4.00 & +3.00 
        & +1.46 & +2.12 & +0.93 
        & +2.15 & +2.30 & +1.38
        \\ \hline
        MoCo-v2-IN        & 87.98 & 77.40 & 74.08 
        & 69.33 & 62.03 & 57.14 
        & 82.66 & 67.38 & 62.42 
        & 79.99 & 68.94 & 64.55 \\
        MoCo-v2-K        & 87.66 & 77.10 & 74.33 
        & 69.24 & 62.42 & 57.86 
        & 80.55 & 65.33 & 60.97
        & 79.15 & 68.28 & 64.39 \\
        DenseCL-IN        & 88.11 & 77.56 & 74.62 
        & 66.56 & 61.42 & 57.08 
        & 83.86 & 68.81 & 64.74 
        & 79.51 & 69.26 & 65.48
        \\ \hline
    \end{tabularx}
    \caption{Camera-lidar fusion based 3D object detection fine-tuned on KITTI validation set. We show the bounding box AP of each class in details. `K' and `IN' indicates pre-trained models are trained on KITTI and ImageNet datset. Best is in \textbf{bold}.}
    \label{tab::kitti_3class_res}
\end{table*}

Given the camera parameter $C$, we can establish the positive correspondences between the downsampled points $(l^{\alpha},\ f^{\alpha})$ in Eq.\ref{eq::extract_features} and the image feature maps $f$ in Eq.\ref{eq::extract_img_features} through the projection matrix. Specifically, we project the downsampled points onto the image plane and sample from the image feature maps to get the corresponding image features $f^{\gamma}$:
\begin{equation}
\label{eq::projection}
f^{\gamma} = f \Bigl< proj(l^{\alpha}, C) \Bigl>,
\end{equation}
where $proj()$ are the resulting 2D coordinates of the projected points. $\bigl<\bigl>$ is the sampling operator. In our work, we utilize bi-linear interpolation to sample features. 

Above Operations lead to positive matches $M_2$. Similar to Eq. \ref{eq::intro_loss}, for each matched pair $(i,\ j)\in M_2$, we calculate the inter-modal contrastive loss as:
\begin{equation}
\label{eq::cross_loss}
L_{inter} = -\sum\limits_{(i,\ j)\in M_2}\log\frac{\exp{(f_{i}^{\alpha} \cdot f_{j}^{\gamma}/\tau)}}{\sum_{(\cdot,\ k)\in M_2}{\exp{(f_{i}^{\alpha} \cdot f_{k}^{\gamma}/\tau)}}}.
\end{equation}

Here, we crop the gradient of $f^{\alpha}$ to avoid influences on Intra-Modal Spatial Perception Module. The contrastive loss imbues the prior spatial knowledge of the lidar feature extractor to the image encoder.

Finally, we train the whole framework in an end-to-end fashion with the total loss:
\begin{equation}
\label{eq::total_loss}
L_{total} = \lambda L_{intra} + \mu L_{inter},
\end{equation}
where $\lambda$, $\mu$ are hyperparameters that balances between the two parts of the loss.

Compared with methods that focus on the indoor RGB-D data~\cite{hou2021pri3d, liu2021learning}, which utilize a U-Net~\cite{ronneberger2015u} shape backbone to align the dense/local feature extracted by point-cloud feature extractor, our method only adopts a standard ResNet-50 encoder to extract the global features. Since the modules apply to global representations, alignment is achieved on the fly. Such characteristic indicates that our method is more general to downstream tasks because there is no more limitation on the downstream network design.

\begin{table*}[htb]
    \centering
    \begin{tabularx}{0.95\linewidth}{@{}c|*{2}{C}|*{2}{C}|*{2}{C}|*{1}{C}c@{}}
    \hline
    \multirow{2}{*}{~~Pre-train~~} &
    \multicolumn{2}{c|}{Vehicle L1/L2} &
    \multicolumn{2}{c|}{Pedestrian L1/L2} &
    \multicolumn{2}{c|}{Cyclist L1/L2} 
    & \multicolumn{2}{c}{Overall L1/L2} 
    \\
        & mAP(\%) & mAPH(\%) & mAP(\%) & mAPH(\%) & mAP(\%) & mAPH(\%) 
        & mAP(\%) & mAPH(\%)     
    \\ \hline
    Scratch        & 65.0/61.0 & 64.6/60.5 
    & 67.6/62.9 & 58.8/54.7 
    & 64.0/61.2 & 61.1/58.5 
    & 65.57/61.53 & 61.75/57.93
    \\
    Ours-W         & 66.5/62.4 & 66.1/62.0 
    & 69.4/64.7 & 60.5/56.3 
    & 64.7/62.3 & 62.3/60.0 
    & 66.92/63.01 & 63.18/59.47
    \\ \hline
    Gain           & +1.5/+1.4 & +1.5/+1.5 
    & +1.8/+1.8 & +1.2/+1.6 
    & +0.7/+1.1 & +1.2/+1.5 
    &+1.35/+1.48 & +1.43/+1.54
    \\
    \hline
    \end{tabularx}
    \caption{Camera-lidar fusion based 3D object detection performance comparison on Waymo validation set. `W' indicates the pre-trained models are trained on Waymo datset.}
    \label{tab::waymo_detail_res}
\end{table*}

\section{Experiments}
\label{sec::experiments}

In this section, we introduce our experimental settings~(Section \ref{subsec::experimental_settings}) and downstream results~(Section \ref{subsec::experimental_results}) in detail. The ablation study is applied to prove the effectiveness of key components in the framework and explore the influence of pre-training data scale in Section~\ref{subsec::ablation_study}.

\subsection{Experimental Settings}
\label{subsec::experimental_settings}

We study pre-training strategies on multi-modal datasets, including the KITTI dataset and the Waymo Open Dataset. Both the datasets contain paired image and point cloud data. All the experiments are based on MMDetection3d~\cite{mmdet3d2020}.


\subsubsection{KITTI Dataset~(K).} 
The KITTI Dataset~\cite{geiger2012we} contains 7481 training images and 7518 test images, both with their corresponding point cloud. To make full use of the dataset, we utilize both the training set and the testing set data to pre-train the model. Note that we filter out the validation set to avoid the information leak.

\subsubsection{Waymo Open Dataset~(W).} 
The Waymo Dataset~\cite{sun2020scalability} contains $\sim$0.15 million training images with corresponding point clouds. The testing set is relatively smaller than the training set. We only use the training set data to pre-train models. Because we need to project lidar points to image plane during the pre-training stage, for simplicity, we filter out points beyond the image field of view (FOV). 

\subsubsection{Pre-training.}
In terms of the backbone settings, we use three set abstraction layers~\cite{qi2017pointnet++, yang20203dssd} to downsample the points and extract point cloud global features. We combine two downsampling strategies to make full use of the data, which are the 3D Euclidean distance sample (D-FPS)~\cite{qi2017pointnet++} and the feature distance sample (F-FPS)~\cite{yang20203dssd}. Following most of the contrastive learning methods, a ResNet-50~\cite{he2016deep} is adopted as the image feature extractor, which can also be replaced by any other image backbone. Compared with the Prior3D~\cite{hou2021pri3d}, our method has fewer constraints on the image backbone and is more general to downstream tasks.

As for contrastive learning settings, the temperature factor $\tau$ in Eq.~\ref{eq::intro_loss} and Eq.\ref{eq::cross_loss} is 0.07. Following~\cite{chen2020simple}, we use the MLP projection head to map the dimension of features to 128-d for either the intra-modal contrastive learning and the inter-modal contrastive learning. We use 4096 matched pairs for faster training~\cite{xie2020pointcontrast}. In addition, we implement the  Moco-v2~\cite{he2020momentum} on both the KITTI dataset and the Waymo dataset to make a fair comparison with other strong unsupervised methods. The data augmentation pipeline of MoCo-v2 consists of random color jittering, random gray-scale conversion, Gaussian blurring, and random horizontal flip. 

\begin{table}[t]
    \centering
    \begin{adjustbox}{width=\linewidth,center}
        \begin{tabular}{c|c|c|c|c|c|c|c}
        \hline
        Pre-train & REL$\downarrow$ & Sq R$\downarrow$ & RMS$\downarrow$ & log$\downarrow$ & $\delta_1$$\uparrow$ & $\delta_2$$\uparrow$ & $\delta_3$$\uparrow$
        \\ \hline
        Scratch           & 0.096 & 0.493 & 3.575 & 0.146 & 0.896 & 0.977 & 0.994
        \\
        Ours-W         & 0.073 & 0.285 & 2.840 & 0.113 & 0.935 & 0.990 & \textbf{0.998}
        \\ \hline
        Gain & -0.023 & -0.208 & -0.735 & -0.033 & +0.039 & +0.013 & +0.004
        \\ 
        \hline
        Super-IN  & 0.068 & 0.247 & 2.712 & 0.104 & 0.946 & \textbf{0.993} & \textbf{0.998}     
        \\ 
        Ours-IN/W      & \textbf{0.067} & \textbf{0.235} & \textbf{2.592} &
        \textbf{0.102} & \textbf{0.949} & \textbf{0.993} & \textbf{0.998}
        \\ \hline
        Gain & -0.001 & -0.012 & -0.120 & -0.002 & +0.003 & 0 & 0
        \\ 
        \hline
        \end{tabular}
    \end{adjustbox}
    \caption{Monocular depth estimation performance comparison on KITTI dataset.  `IN/W' indicates the double fine-tuning pre-trained models on ImageNet and Waymo.}
    \label{tab::monodepth}
\end{table}

\subsection{Experimental Results}
\label{subsec::experimental_results}

\subsubsection{3D Object Detection.}
We evaluate the pre-trained models by fine-tuning on the target 3D-related tasks. For early-fusion based 3D object detection, we utilize two challenging and popular datasets, \textit{i.e.}, KITTI and Waymo. We fine-tune the pre-trained models with the state-of-the-art algorithms: MVX-Net~\cite{sindagi2019mvx} with Moca~\cite{zhang2020multi} on KITTI. For the Waymo dataset, we find that the Moca can not improve the MVX-Net performance. Therefore, we choose to only apply the MVX-Net as our default protocol, which is still a strong baseline of early-fusion based methods. When evaluating, we use the standard 2$\times$ schedule, which is more effective on the waymo dataset. 


The KITTI 3D object detection performance comparison is shown in Tab.~\ref{tab::kitti_3class_res}. We utilize the scratch one as our baseline method. Compared with it, our method achieves the significant 2.3\% moderate overall $AP_{3D}$ gains. Furthermore, 1.38\% gains on the hard overall $AP_{3D}$ shows that our pre-training method improves the localization accuracy. To further dig into the effectiveness of the pre-training methods, we report the per-class comparison results. Our fine-tuned model can localize small objects more accurately, which is reflected in significant 4\% gains on pedestrian moderate $AP_{3D}$, compared with other classes. More results compared with other state-of-the-art counterpart pre-training methods can be found in Fig.~\ref{fig::pilotstudy} and the appendix:.

\begin{figure}[t]
    \centering
    \includegraphics[width=2.9in]{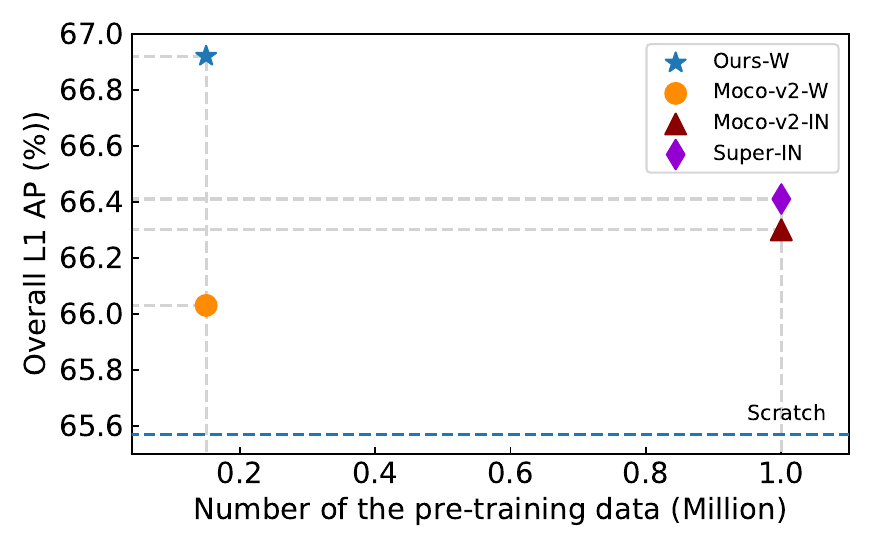}
    \caption{Fusion-based 3D object detection performance comparison on Waymo dataset. SimIPU achieves comparable results with limited multi-modal pre-training data.}
    \label{fig::waymores}
\end{figure}

In Fig.~\ref{fig::waymores}, we report the 3D object detection results on Waymo Dataset. Waymo dataset is much larger than KITTI. Although the performance difference between pre-trained and scratch models is not obvious for larger dataset~\cite{xie2020pointcontrast}, our method still achieves a slight improvement even using limited data. Compared with training from scratch, our method achieves a relatively significant 1.35\% mAP improvement. Similar to the results on KITTI, our fine-tuned model has the capability to localize small objects more accurately. We provide more results in the appendix.

\begin{table}[t]
    \centering
    \begin{adjustbox}{width=\linewidth,center}
        \begin{tabular}{c|c|c|c|c|c|c|c}
        \hline
        Pre-train & AP(\%) & ATE & ASE & AOE & AVE & AAE & NDS(\%)
        \\ \hline
        Scratch & 17.90 & 0.92 & 0.30 & 0.84 & 1.33 & 0.19 & 26.27
        \\
        Ours-W   & 26.18 & 0.84 & 0.27 & 0.67 & 1.31 & 0.17 & 33.50
        \\ \hline
        Gain & +8.28 & - & - & - & - & - & +7.23
        \\ 
        \hline
        Super-IN & 27.71 & 0.83 & 0.26 & 0.59 & 1.34 & 0.16 & 35.23
        \\
        Ours-IN/W  & \textbf{28.36} & 0.82 & 0.26 & 0.62 & 1.33 & 0.16 & \textbf{35.36}
        \\ \hline
        Gain & +0.65 & - & - & - & - & - & +0.13
        \\
        \hline
        \end{tabular}
    \end{adjustbox}
    \caption{Monocular 3D object detection performance comparison on Nuscenes dataset.}
    \label{tab::mono3d}
\end{table}

\subsubsection{KITTI Monocular Depth Estimation.}
As for the monocular depth estimation, we design a simple yet strong baseline to evaluate the model performance. We provide more information of the baseline in the appendix. We evaluate the effectiveness of our method by fine-tuning pre-trained the model on the KITTI Eigen split~\cite{eigen2015predicting}. All settings are the same when doing fine-tuning to make a fair comparison. 


In Tab.~\ref{tab::monodepth}, we report the KITTI monocular depth estimation performance. Compared with training from scratch, our methods achieve great improvement on all of the metrics. However, we can only obtain $\sim$0.15M~(Million) data to pre-train our model, which is extremely smaller than the ImageNet supervised pre-training methods that can utilize $\sim$1M labeled data to pre-train models, therefore, there is a small performance gap between the results of Ours-W over Super-IN. To fill this gap, we propose a simple double fine-tuning strategy. We load the pre-trained Super-IN model at the beginning of our multi-modal contrastive pre-training stage and double fine-tune the model on the downstream task. As a result, our method can learn useful spatial-aware visual representations and preserve part of the semantic representations, which leads to further performance gains, especially for the 0.12m~(4.4\%) improvement on RMS. 

\subsubsection{Nuscenes Monocular 3D Object Detection.}

In terms of the monocular 3D object detection, FCOS3D~\cite{wang2021fcos3d} is fine-tuned on Nuscenes training set and evaluated on Nuscenes validation set. For  simplicity, we only replace the image encoder ResNet-101 in the default config of FCOS3D with ResNet-50. When evaluating, we use the standard 1$\times$ schedule. The batch size is set to 8. Synchronized batch normalization is used. All settings are default.

We report the performance of Nuscenes monocular 3D object detection in Tab.~\ref{tab::mono3d}. Our methods achieve significant improvement on all of the metrics compared with training from scratch. Applying the same double fine-tuning strategy, our method further boosts the performance of the Super-IN model by 0.65\% mAP.

\subsection{Ablation Study}
\label{subsec::ablation_study}

\subsubsection{Intra-Modal Module.}
Viewing the whole framework, our method can be treated as an indirect method for image feature extractors to learn a spatial-aware visual representation. Knowledge transferred from the point cloud encoder is crucial and determines the effectiveness of our method. We design the first ablation study to prove if the intra-modal contrastive learning module does have learned a useful representation. The results shown in the first block of Tab.~\ref{tab::ablation_intro_kitti} indicate that the intra-modal branch is essential in the multi-modal contrastive learning framework. With the intra-modal branch, the point cloud feature extractor can learn a spatial-aware visual representation and transfer it to the image encoder via inter-modal contrastive learning, which can further boost the performance of 3D-related tasks. 


\begin{table}[t]
    \centering
    \begin{adjustbox}{width=1\linewidth,center}
        \begin{tabular}{c|ccc}
        \hline
        \multirow{2}{*}{Pre-train}  & \multicolumn{3}{c}{Overall $\text{AP}_{3D}$(\%)} \\
        & Easy     & Mod.     & Hard 
        \\ \hline
        Scratch (Equivalent to $\lambda = 1$, $\mu$ = 0) & 79.17 & 68.58 & 64.81 \\
        SimIPU w/o intra-module ($\lambda = 0$, $\mu$ = 1) & 79.36 & 69.19 & 65.17 \\
        SimIPU w/o inter-module ($\lambda = 1$, $\mu$ = 0) & 79.17 & 68.58 & 64.81 \\
        SimIPU ($\lambda = 1$, $\mu = 1$) & \textbf{81.32} & \textbf{70.88} & \textbf{66.19}
        \\ \hline
        Greedy Assignment       & 78.92 & 68.20 & 64.72 \\
        Hungarian Algorithm     & \textbf{81.32} & \textbf{70.88} & \textbf{66.19} \\
        \hline
        \end{tabular}
    \end{adjustbox}
    \caption{Ablation study on intra-modal contrastive learning module and matching strategies. We report 3D object detection performance results on KITTI validation set.}
    \label{tab::ablation_intro_kitti}
\end{table}

\subsubsection{Matching Algorithm.}
We use the Hungarian Algorithm as our default assignment method, which can achieve the optimal global solution for the positive pair association. In this ablation study, we compare it with another commonly used matching algorithm: Greedy Assignment. During the matching process, this algorithm takes the nearest optimal options and repeats them. The ablation experimental results are shown in the last block of Tab.~\ref{tab::ablation_intro_kitti}. Greedy assignment hampers the downstream performance. The reason may be caused by poor quality matches. It hurts the effectiveness of intra-modal contrastive learning, which is essential to learn useful representations for downstream tasks.

\subsubsection{Number of Pre-training Data.}
\begin{figure}[t]
    \centering
    \includegraphics[width=3.3in]{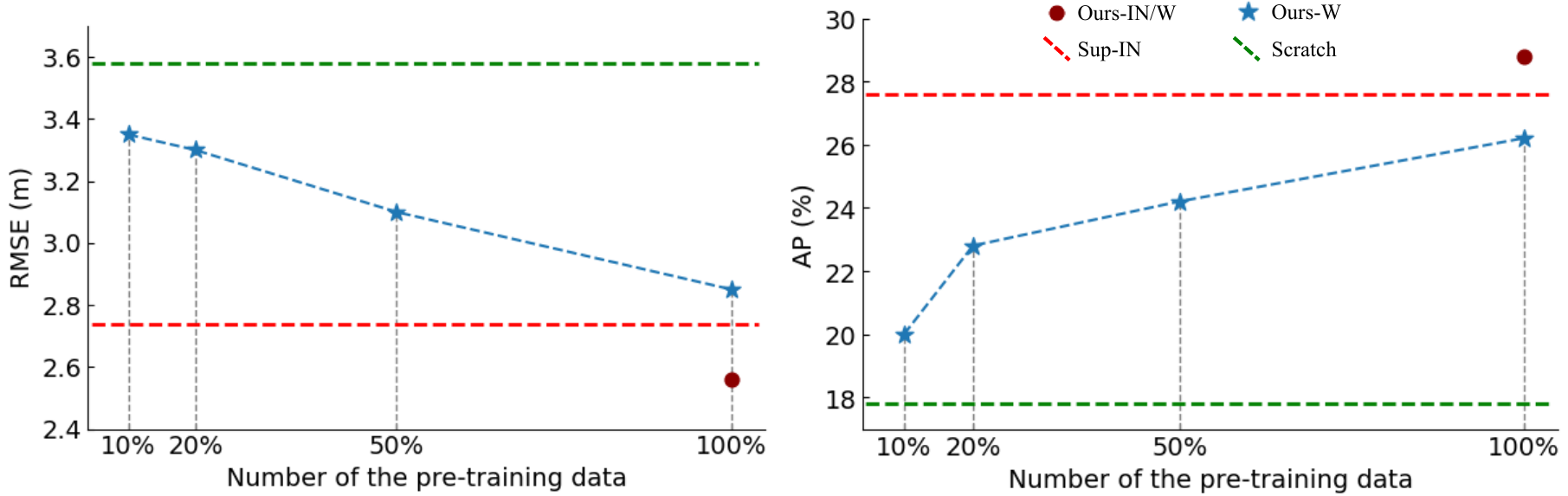}
    \caption{Ablation study on different pre-training data scale. We report the RMSE values on KITTI monocular depth estimation (left) and the AP results on Nuscenes monocular 3D object detection (right).}
    \label{fig::datascale}
\end{figure}

In this ablation experiment, we use different amounts of training data to pre-train models on the Waymo dataset and fine-tune them on both the monocular depth estimation and the monocular 3D object detection task, to explore the influence of the number of pre-training data on downstream tasks. Results are shown in Fig.~\ref{fig::datascale}. One can easily observe that the performance of downstream tasks is significantly improved by the increase of pre-training data. It is in line with our intuition: a larger scale of pre-training data will further boost the performance of downstream tasks. Note that we only use $\sim$0.15M unlabeled data to pre-train our model. Compared with Super-IN, which utilizes $\sim$1M labeled data to pre-train models, our method undoubtedly shows significant competitiveness. The curves of the experimental results can thus be suggested that our method will easily surpass the Super-IN with more pre-training data. In addition, by a simple double fine-tuning strategy, our method can further boost the performance of baseline models, which indicates the friendly compatibility and generalization of our method.

\section{Conclusion}
\label{sec::conclusion}

In this paper, we propose SimIPU, a simple yet effective 2D Image and 3D Point cloud Unsupervised pre-training method, and develop a multi-modal contrastive learning framework to learn spatial-aware visual representation for 3D-related tasks in the outdoor environment. This method fills the blank of pre-training methods for outdoor multi-modal datasets and achieves significant performance gains on different 3D-related downstream tasks, including fusion-based 3D object detection, monocular depth estimation, and monocular 3D object detection, with a limited number of multi-modal pre-training data. However, an inadequate of this method is that it only focuses on spatial-aware visual representation while ignores the semantic information, but even notwithstanding this limitation, our approach still shows great generalization and effectiveness on downstream tasks. In the long term, associating both the spatial and the semantic information would be a fruitful area for further work, and we will dig into more effective methods to achieve this purpose. We hope our work will encourage more research on visual representation learning for a suitable design of the cross-modal pre-training paradigm.

\section{Acknowledgments}
The research was supported by the National Natural Science Foundation of China (61971165, 61922027), in part by the Fundamental Research Funds for the Central Universities (FRFCU5710050119). The authors would like to thank Jinghuai Zhang from the Computer Science Department of Duke University for helpful discussions on topics related to this work.

\bibliography{aaai22}

\appendix

\section{KITTI 3D Object Detection Results}
As shown in Tab.~\ref{sup_tab::kitti_3class_res}, we report more results on KITTI 3D Object Detection. In pilot study, we utilize different pre-training models, such as supervised 2D object detection pre-trained ones, supervised classification pre-trained ones, and other state-of-the-art counterpart unsupervised pre-trained ones, to initialize the image backbone. Our method achieves significant improvement that indicates the effectiveness.

\begin{table*}[h]
    \centering
    \footnotesize
    {
    \begin{tabular}{c|ccc|ccc|ccc|ccc}
    \toprule
    \multirow{2}{*}{Pre-train}  & \multicolumn{3}{c|}{Car $\text{AP}_{3D}$(\%)} & \multicolumn{3}{c|}{Pedestrian $\text{AP}_{3D}$(\%)}  & \multicolumn{3}{c|}{Cyclist $\text{AP}_{3D}$(\%)}
    & \multicolumn{3}{c}{Overall $\text{AP}_{3D}$(\%)}
    \\
    & Easy     & Mod.     & Hard   & Easy     & Mod.     & Hard   & Easy     & Mod.     & Hard   
    & Easy     & Mod.     & Hard 
    \\ \midrule
    Scratch        & 86.18 & 76.57 & 74.08 & 67.95 & 62.18 & 57.24 & 83.37 & 66.99 & 63.11 
    & 79.17 & 68.58 & 64.81 \\
    MoCo-v2-IN        & 87.98 & 77.40 & 74.08 & 69.33 & 62.03 & 57.14 & 82.66 & 67.38 & 62.42 
    & 79.99 & 68.94 & 64.55 \\
    MoCo-v2-K        & 87.66 & 77.10 & 74.33 & 69.24 & 62.42 & 57.86 & 80.55 & 65.33 & 60.97
    & 79.15 & 68.28 & 64.39 \\
    DenseCL-IN        & 88.11 & \textbf{77.56} & 74.62 & 66.56 & 61.42 & 57.08 & 83.86 & 68.81 & \textbf{64.74} 
    & 79.51 & 69.26 & 65.48 \\
    DenseCL-Co        & 88.29 & 77.46 & 74.47 & 65.53 & 59.42 & 53.33 & 83.59 & 68.86 & 63.72
    & 79.14 & 68.58 & 63.84 \\
    Det-K        & 88.23 & 77.54 & 74.82 & 65.94 & 60.04 & 55.46 & 80.82 & 65.74 & 61.01
    & 78.33 & 67.77 & 63.76 \\
    Det-CS        & 87.68 & 77.29 & 74.83 & 67.44 & 61.04 & 56.14 & 83.26 & 67.39 & 62.70
    & 79.46 & 68.73 & 64.52 \\
    Super-IN        & \textbf{88.38} & 77.49 & \textbf{74.94} & 67.20 & 60.06 & 55.77 & \textbf{85.07} & 67.47 & 62.60
    & 80.21 & 68.34 & 64.44 \\
    Ours-K         & 87.87 & 77.36 & 74.30 & \textbf{71.25} & \textbf{66.18} & \textbf{60.24} & 84.83 & \textbf{69.11} & 64.04 
    & \textbf{81.32} & \textbf{70.88} & \textbf{66.19}
    \\
    \bottomrule
    \end{tabular}
    }
    \caption{Camera-lidar fusion based 3D object detection fine-tuned on KITTI validation set. `K', `CS', `Co' and `IN' indicates pre-trained models are trained on KITTI, Cityscapes, COCO and ImageNet dataset. `Det' represents the 2D detection task.}
    \label{sup_tab::kitti_3class_res}
\end{table*}

\begin{table*}[h]
    \centering
    \footnotesize
    {
    \begin{tabular}{c|cc|cc|cc|cc}
    \toprule
    \multirow{2}{*}{Pre-train} &
    \multicolumn{2}{c|}{Vehicle L1/L2} &
    \multicolumn{2}{c|}{Pedestrian L1/L2} &
    \multicolumn{2}{c|}{Cyclist L1/L2} 
    & \multicolumn{2}{c}{Overall L1/L2} 
    \\
        & mAP(\%) & mAPH(\%) & mAP(\%) & mAPH(\%) & mAP(\%) & mAPH(\%) 
        & mAP(\%) & mAPH(\%)     
    \\ \midrule
    Scratch        & 65.0/61.0 & 64.6/60.5 
    & 67.6/62.9 & 58.8/54.7 
    & 64.0/61.2 & 61.1/58.5 
    & 65.57/61.53 & 61.75/57.93
    \\
    Super-IN& 66.0/61.9 & 65.5/61.5 
    & \textbf{69.5}/\textbf{64.8} & \textbf{60.9}/\textbf{56.7} 
    & 63.7/61.3 & 60.4/58.1 
    & 66.41/62.66 & 62.26/58.76
    \\
    MoCo-v2-IN & 66.1/62.0 & 65.6/61.6 
    & 69.0/64.3 & 60.0/55.8 
    & 63.8/61.0 & 60.9/58.3 
    & 66.30/62.43 & 62.16/58.56
    \\
    MoCo-v2-W  & 65.5/61.5 & 65.1/60.7 
    & 68.6/63.9 & 59.1/54.9 
    & 64.0/61.2 & 61.1/58.5 
    & 66.03/62.16 & 61.76/58.03
    \\
    Ours-W         & \textbf{66.5}/\textbf{62.4} & \textbf{66.1}/\textbf{62.0} 
    & 69.4/64.7 & 60.5/56.3 
    & \textbf{64.7}/\textbf{62.3} & \textbf{62.3}/\textbf{60.0} 
    & \textbf{66.92}/\textbf{63.01} & \textbf{63.18}/\textbf{59.47}
    \\
    \bottomrule
    \end{tabular}
    }
    \caption{Camera-lidar fusion based 3D object detection performance comparison on Waymo validation set. `W' indicates pre-trained models are trained on Waymo datset.}
    \label{sup_tab::waymo_3class_res}
\end{table*}

\begin{table}[h]
    \centering
    \footnotesize
    \begin{tabular}{c|c|c|c|c}
    \toprule
    Method & Backbone & REL$\downarrow$ & RMS$\downarrow$ & $\delta_1$$\uparrow$ 
    \\ 
    \midrule
    Our Baseline &   ResNet-50  & 0.068 & 2.712 & 0.946
    \\
    Adabins & Efficient-B5   & 0.058 & 2.360 & 0.964
    \\
    \bottomrule
    \end{tabular}
    \caption{Depth Estimation Baseline Comparison.}
    \label{sup_tab::depth}
\end{table}

\section{Waymo 3D Object Detection Results}
As shown in Tab.~\ref{sup_tab::waymo_3class_res}, we report more results on Waymo 3D Object Detection. Waymo dataset is much larger than KITTI. Although the performance difference between pre-trained and scratch models is not obvious for larger dataset~\cite{xie2020pointcontrast}, our method still achieves a promising improvement even with limited data.

\section{Depth Estimation Baseline Design}
In this paper, we design a simple depth estimation baseline to evaluate the effectiveness of pre-trained models. The whole UNet-shape framework contains two components: 1) a standard ResNet-50 encoder and 2) an upsampling decoder~\cite{bhat2021adabins}. Skip connections are applied to consolidate feature maps from higher resolutions. Therefore, it outputs depth maps with half the spatial resolution. After upsamping the prediction depth maps to the same resolution as the input images, we utilize the Scale-Invariant loss (SI) introduced by Eigen et al.~\cite{eigen2015predicting} to train the model:
\begin{equation}
L = \alpha \sqrt{\frac{1}{T}\sum_{i}g_{i}^2 + \frac{\lambda}{T}(\sum_{i}g_{i})^2}
\end{equation}
\noindent where $g_i = log \hat d_i - log d_i$ and the ground truth depth $d_i$ and $T$ denotes the number of pixels having valid ground truth values. We use $\lambda = 0.85$ and $\alpha = 10$ for all our experiments, which is the same as the Adabins~\cite{bhat2021adabins}.

This simple baseline is strong enough to evaluate the effectiveness of our pre-training method. The comparison with the state-of-the-art depth estimation algorithm is shown in Tab.~\ref{sup_tab::depth}, where both the encoders are pre-trained on ImageNet for fair comparison.

\section{Pre-training Setting Details}
During pre-training, we use the hybrid optimize strategy~\cite{zhang2020multi} to train the whole framework in an end-to-end manner. For the image branch, we use SGD as our optimizer. The momentum is set to 0.9, weight decay is 0.0001 and learning rate is 0.03. For lidar branch, we use AdamW as our optimizer. The $\beta$ is set to (0.95, 0.99), weight decay is 0.01 and learning rate is 0.001. The loss weight in total loss is experimentally set to 1, because we find the convergence of lidar branch is faster than image~\cite{hou2021pri3d}, by inter-modal contrastive learning, it can gradually transfer the spatial aware representation to image modality. We train our model on KITTI dataset for 100 epoches and Waymo dataset for 10 epoches with a batch size of 4 on 8 NVIDIA TITAN V100 GPUs. It takes around 24 hours and 48 hours to pre-train models on KITTI and Waymo Dataset, respectively.

\section{Single-Modality Downstream Task Results}
Limited by pre-training data scale, our method can not directly beat the Super-IN when fine-tuning on the single-modality downstream tasks, \textsl{i.e.,} monocular depth estimation and monocular 3D object detection. As shown in the ablation study, with more data, our method has a high probability of surpassing the ImageNet supervised ones. To further exert the advantages of multi-modal pre-training, we propose a simple double fine-tuning strategy. In this way, although the data scale is limited, we can fine-tune a strong baseline model~(Super-IN) with our method and further improve the model performance on these single-modality downstream tasks. To prove the effectiveness of our method, we need to compare it with other unsupervised pre-training methods with the same double fine-tuning strategy. However, we find that double fine-tuned models with other pre-training methods (MoCo-v2) \textbf{cannot} converge on these downstream tasks, which indicates that the representations learned from different pre-training strategies~(supervised pre-training and unsupervised pre-training) designed on 2D image planes maybe not compatible. Such phenomena shows the effectiveness and generalization of our method.

\end{document}